\documentclass[runningheads]{llncs}
\usepackage{graphicx}
\usepackage{amsmath,amssymb} 

\usepackage{color}
\usepackage{epsfig}
\usepackage{graphicx}
\usepackage{bbm}
\usepackage[bold]{hhtensor}

\DeclareMathOperator*{\argmax}{argmax} 

\usepackage[misc]{ifsym}
\begin{document}
\title{Zero-Annotation Object Detection with Web Knowledge Transfer} 

\titlerunning{Zero-Annotation Object Detection with Web Knowledge Transfer}
%
\author{Qingyi Tao\inst{1,2}\textsuperscript{(\Letter)} \and
Hao Yang\inst{3}\thanks{This work was done when Hao Yang was at NTU, Singapore} \and
Jianfei Cai\inst{1}}
%
\authorrunning{Q. Tao, H. Yang and J. Cai}
%

\institute{Nanyang Technological University, Singapore\\
\email{qtao002@e.ntu.edu.sg, asjfcai@ntu.edu.sg}\\
\and
NVIDIA AI Technology Center
\and
Amazon Rekognition
\email{lancelot365@gmail.com}
}
\maketitle              
\begin{abstract}
Object detection is one of the major problems in computer vision, and has been extensively studied. Most of the existing detection works rely on labor-intensive supervision, such as ground truth bounding boxes of objects or at least image-level annotations. On the contrary, we propose an object detection method that does not require any form of human annotation on target tasks, by exploiting freely available web images. In order to facilitate effective knowledge transfer from web images, we introduce a multi-instance multi-label domain adaption learning framework with two key innovations. First of all, we propose an instance-level adversarial domain adaptation network with attention on foreground objects to transfer the object appearances from web domain to target domain. Second, to preserve the class-specific semantic structure of transferred object features, we propose a simultaneous transfer mechanism to transfer the supervision across domains through pseudo strong label generation. With our end-to-end framework that simultaneously learns a weakly supervised detector and transfers knowledge across domains, we achieved significant improvements over baseline methods on the benchmark datasets. 
\end{abstract}
\section{Introduction}

In recent years, with the advances of deep convolutional neural networks (DCNN), object detection tasks have attracted significant attention and have achieved great improvements in performance and efficiency. State-of-the-art works such as Faster R-CNN~\cite{ren2015faster}, SSD~\cite{liu2016ssd}, FPN~\cite{lin2017feature} achieve high accuracy but require labour-intensive bounding box annotations for training. To alleviate the large labour cost for annotating ground truth bounding boxes, weakly supervised object detection methods that only rely on image-level human annotations have also been extensively studied~\cite{bilen2014weakly,bilen2015weakly,bilen2016weakly,cinbis2017weakly,jie2017deep,kantorov2016contextlocnet,tang2017multiple,kumar2016track}. However, for large-scale multi-object detection problem, even annotating just image-level labels could deem to be too expensive. This motivates us to develop an object detection method with no human annotations involved. Our basic idea is to transfer knowledge from free web resources to the target tasks.

\begin{figure}[t]
\begin{center}
 \includegraphics[width=0.6\linewidth]{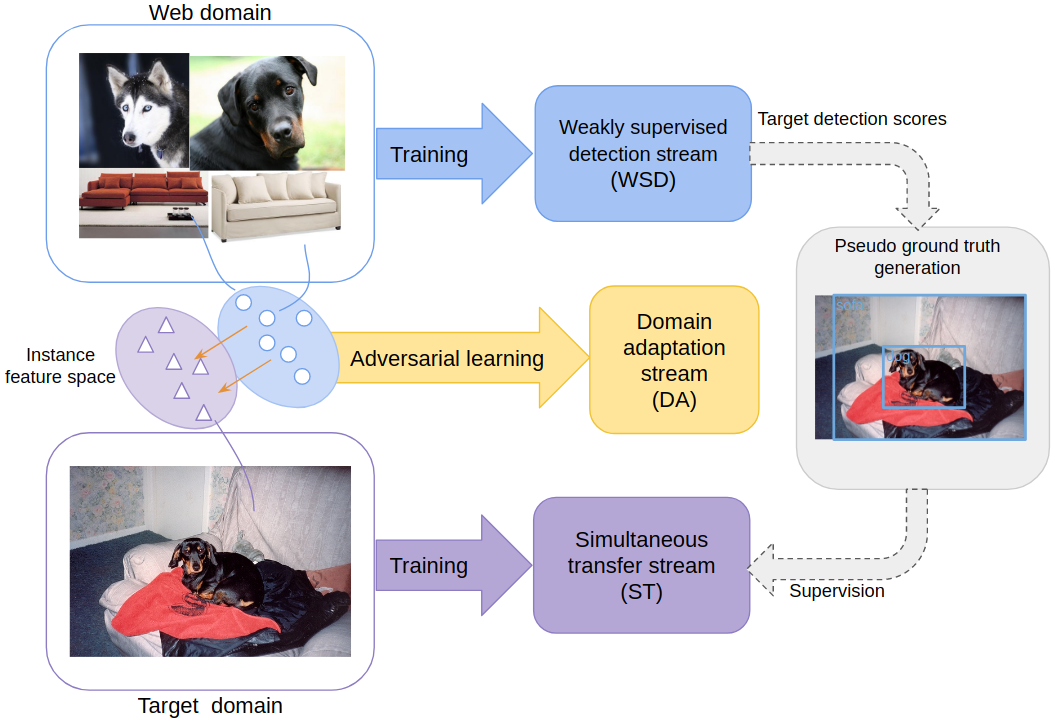}
\end{center}
  \caption{Overall idea of object detection without human annotations. First of all, we mine freely available web images through automatic retrieval with respect to a given set of object categories. Our framework then facilitates knowledge transfer from these web images to the target task using a multi-stream network with three major components: 1) a weakly supervised detection stream (WSD) to train the detection model from web images; 2) an instance-level domain adaptation (DA) stream to minimize the feature discrepancy across domains at instance-level feature space; 3) a simultaneous transfer (ST) stream that learns to discriminate unsupervised target examples by transferring supervision from web detection model. These three streams are trained simultaneously to effectively transfer the learning of web images to the target task.}\label{fig:overall_idea}
\end{figure}

With the similar motivation, zero-shot learning (ZSL) problem has been proposed for unsupervised learning. Many works ~\cite{lampert2009learning,lampert2014attribute,ferrari2008learning,pennington2014glove,fu2015zero,al2017automatic} have been proposed to utilize side information such as attributes, Wikipedia or WordNet to jointly encode semantic space and image feature space for solving zero-shot recognition problems. However, although textual side information could help zero-shot object recognition with exploiting the intrinsic semantic relations between categories, it is hard to learn a class-specific object detector that can accurately differentiate objects from the background as well as different objects with just semantic descriptions. In contrast, our direction is to exploit freely available web images as a much stronger side information to solve the object detection problem without human annotations, considering that there are huge amount of image resources from the web and plenty of works studying the automatic collection of these web imagery resources~\cite{li2010optimol,chen2013neil,xia2014well,tao2017exploiting}.

One baseline approach for learning detectors with web images is to simply use the web images and their image ``labels" (essentially the pre-defined labels used as search phrases to retrieve the images) to train a web object detector using some weakly supervised detection (WSD) methods and apply them on target images. This naive learning scheme is referred as webly supervised learning in previous works ~\cite{divvala2014learning,chen2015webly}. However, directly applying the web models to the target data produces poor results. The major reason is that it ignores the domain discrepancies between web images and target images. As shown in Fig.~\ref{fig:overall_idea}, web images from image search engines are mostly studio-shot images, which are simple, clear and unblocked. In contrast, the target images (e.g. Pascal VOC images) usually contain multiple objects of different classes that are often occluded with cluttered scenes. Hence it is necessary to properly transfer the models learned from web images to the target images.

To address this domain discrepancy problem, we need to adapt the source (i.e. web domain) and target domain object appearances, for which unsupervised domain adaptation is the common way~\cite{tzeng2015simultaneous,ganin2015unsupervised,long2016deep,tzeng2017adversarial,bousmalis2017unsupervised}. Although many unsupervised domain adaptation methods have been proposed, they all focus on image-level domain adaptation for image classification problems. What we consider here is the domain adaptation at instance level (i.e. object proposal level), which is non-trivial to solve. Inspired by the recent adversarial domain adaptation works ~\cite{tzeng2015simultaneous,ganin2015unsupervised,tzeng2017adversarial}, we propose an instance-level adversarial domain adaptation network to reduce the domain discrepancies particularly at instance level. Our adversarial domain adaptation network includes a domain discriminator that differentiates object features from web domain and target domain, and a feature generator that projects source and target objects to the same manifold in the feature space so that the discriminator can no longer tell their differences. 

In addition, we introduce an innovative component in our domain adaptation network: attention on foreground objects. As weakly supervised detection is essentially a multi-instance multi-label learning problem, each image actually is a bag of instances, where each instance corresponds to a bounding box proposal. Equally treating all proposals in each image when training adversarial domain adaptation network will lead to sub-par results, as we care more about proposals containing objects than proposals that are largely background. Therefore, we introduce an attention mechanism to emphasize the transfer of object proposals and suppress the transfer of background proposals. 

However, the introduced instance-level domain adaptation network brings in a side effect, i.e. the feature generator is likely to ignore the semantic structure of different object classes, since there is no class-specific constraint. As a result, it not only brings features from different domains together to the same manifold, but also mixes up the sub-manifolds from different classes. For example, the ``cow" from web domain will be confused with the ``sheep" from target domain through the domain adaptation. To address this issue, we further introduce simultaneous learning towards class-specific pseudo labels to preserve the semantic structure during the domain adaptation. This component compensates the side effect of the domain adaptation component so that the domain shift will be guided in a class-specific manner. In this way, our overall architecture including the web object detector, the domain adaptation component and the simultaneous transfer component significantly boosts up the object detection results on unsupervised target data.

We would like to highlight that the rationale of studying this problem lies in that such detector can be trained without any human labour and therefore the whole process could be fully automated. Different from fully supervised and weakly supervised object detection, our object detector allows the training of the detection models to be highly scalable in term of categories.
For example, in the Pascal VOC dataset, if we want to add the object class ``keyboard", which exists in some of the images but is not annotated, we need to re-annotate all the images in the training data by providing respective labels at bounding box level (for supervised detector) or image level (for weakly supervised detector). Another example is that if we want to further break down the ``bird" class into multiple classes such as ``parrot", ``goose", ``hawk" and etc, we also need to revise the annotations for all images containing ``bird" objects. In contrast, our solution can automatically search the web and progressively transfer the web knowledge to learn the detector without any human intervention or any modification in the target domain dataset. The training of such detector can be a completely self-taught process. Hence, we think this problem is highly meaningful and worth to be studied.

Overall, the main contributions of our work can be summarized as follows:
\begin{itemize}
  \item We propose a new problem of knowledge transfer in object detection for \textbf{\textit{unsupervised}} data, which enables learning an object detector from free web images and alleviates any forms of human annotations for target domain. By studying this problem, the learning of object detectors can be fully automated and highly scalable with categories. 
  \item We propose an \textbf{\textit{instance-level}} domain adaptation method to transfer web knowledge to unsupervised target dataset. The proposed domain adaptation framework includes: 1) an instance-level adversarial domain adaptation network with attention on foreground objects; 2) 
  a simultaneous transfer stream to preserve the semantic structure of classes by transferring the pseudo labels obtained from the web domain detector to the target domain detector. 
  \item Our method significantly reduces the gap between unsupervised object detection (i.e. train a detector using only web images and then directly apply it on target images) and the upper bound (i.e. train a detector using image-level labels of target data) by 3.6\% in detection mAP.
\end{itemize}


\section{Related Works}\label{sec:related_work}
Our work is related to a few computer vision and machine learning areas. We will review these related topics in this section.

\textbf{Weakly supervised object detection:} Recent works on weakly supervised object detection aim to reduce the intensive human labour cost by using only image-level labels instead of bounding box annotations \cite{bilen2014weakly,bilen2015weakly,cinbis2017weakly}. They are more cost-effective than the fully supervised object detection methods since image-level labels are easier to obtain compared with the bounding box annotations. These works formulate the weakly supervised object detection task as a multi-instance learning (MIL) problem in which the model will be learned alternatively to recognize the object categories and to find the object locations of each category. The recent work~\cite{bilen2016weakly} is the first one introducing an end-to-end network with two separate branches for object recognition and localization respectively. Later, \cite{kantorov2016contextlocnet} introduced context information to the weakly supervised detection network in the localization stream. \cite{tang2017multiple} proposed online classifier refinement to refine the instance classification based on image-level labels. 

Our work is related to these works as the web data will be trained in a weakly supervised way with their weak labels. In this paper, we use WSDDN in \cite{bilen2016weakly} as the base model for our work.

\begin{figure*}[t]
\begin{center}
\includegraphics[width=0.8\linewidth]{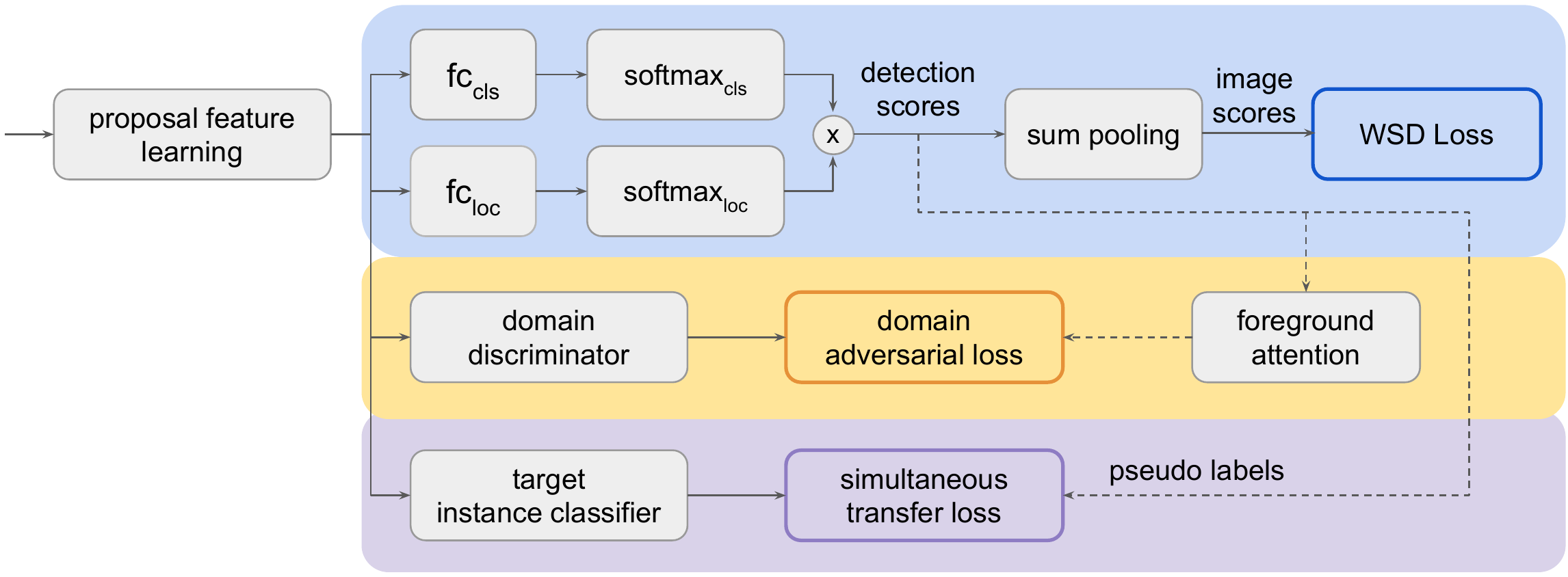}
\end{center}
\caption{The proposed network branches into three streams after the proposal feature learning layers. The first stream (in blue) is the weakly supervised detection (WSD) network which is further divided into recognition and localization streams. The middle stream (in yellow) is the instance-level domain adaptation (DA) stream that optimizes an adversarial loss to enforce domain invariant feature representation. The last stream (in purple) is the simultaneous transfer (ST) stream to preserve semantic structure of target data with pseudo labels.}
\label{fig:network_architecture}
\end{figure*}

\textbf{Learning from web data:} Web data is a free source of training samples that can be collected automatically for various tasks \cite{chen2015webly,divvala2014learning,xu2015augmenting,sultani2016if}. Previous works \cite{chen2015webly,divvala2014learning,xu2015augmenting} study the web data collection approaches and further evaluate their data collection methods by training those web data for different tasks. They focus on reducing the effects of noises from web images and thereby construct robust and clean web datasets. While learning for the target tasks, these prior works simply treat the web dataset as the substitute of the training dataset in the target task without considering the domain shifts between web data and target data, which is similar as our baseline approach.  Apart from that, web data are often used as complementary data to improve the training of target dataset. In \cite{wei2017stc}, web images are used to produce pseudo masks for pre-training the semantic segmentation network. In \cite{wang2017learning}, an object interaction dataset with web images is created to facilitate the semantic segmentation task as additional data. In their approaches, the image-level labels (in \cite{wei2017stc}) or pixel-level ground truth masks (in \cite{wang2017learning}) of target images are required and web images are utilized as additional knowledge to improve the segmentation model performance. In our work, we attempt to solve the detection problem using only the web images without any forms of annotations from target dataset.

\textbf{Domain adaptation:} Our work is also closely related to the domain adaptation works \cite{tzeng2015simultaneous,ganin2015unsupervised,tzeng2017adversarial,bousmalis2017unsupervised,yang2018shuffle}. \cite{ganin2015unsupervised} introduced the domain adversarial training of neural networks. The domain adaptation is achieved by introducing a domain classifier to classify features to their corresponding domains and applying a gradient reversal layer between the feature extractor and the domain classifier. With this reversal layer, when the domain classifier learns to distinguish the features from different domains, the feature extractor learns in the reverse way to make the feature distributions as indistinguishable as possible. Hence, this domain adversarial training can result in a domain-invariant feature representation. \cite{tzeng2015simultaneous} also uses a similar method for domain transfer in image classification task. In \cite{tzeng2015simultaneous}, a domain classification loss and a domain confusion loss influence the training in an adversarial manner. They also added a soft label layer while learning the source examples in order to transfer correlations between classes to the target examples. Later, \cite{tzeng2017adversarial} proposed to untie the weight sharing between two domains. These previous works have validated the effectiveness of the adversarial domain adaptation methods in the image classification problem. In our work, we follow the principles of the end-to-end adversarial methods but for our zero-annotation detection task with the domain transfer of proposal-level features to reduce the domain mismatch between web data and target data.

\section{Problem Definition and Notations}

In this section, we formally define our problem of zero-annotation object detection with web knowledge transfer. Essentially, we define this problem as an unsupervised multi-instance multi-label domain adaptation problem. Specifically, we consider two domains, the web domain $D^w$ representing web images and target domain $D^t$ representing target tasks (e.g. Pascal VOC and MS COCO). The source data $\{X_j^w,{y}_{j}\}_{j=1}^{n_w}$ is sampled from $D^w$, where $X_j^w$ is the $j$-th image, ${y}_{j} \in \mathbbm{R}^C$ sampled from label space ${Y}$ is the corresponding $C$ dimensional binary label vector and $n_w$ is the number of source images. For object detection problems, it is natural to decompose each image to a bag of instances, i.e., object proposals, through dense sampling or objectness techniques. Thus,  $X_j^w$ can be represented as $X_j^w = \{x_i^{w,j}\}_{i=1}^{m_j^{w}}$, where $x_i^{w,j}$ is the $i$-th proposal in $X_j^w$ and $m_j^{w}$ is the number of total proposals of $X_j^w$. Similarly, the target data sampled from $D^t$ can be denoted as $\{X_j^t\}_{j=1}^{n_t}$, and $X_j^t = \{x_i^{t,j}\}_{i=1}^{m_j^{t}}$. Note that since we do not have annotations for target data, effective knowledge transfer from the web domain is necessary.

Traditional domain adaptation methods usually optimize an objective function $f:X^w, X^t \rightarrow {Y}$, which jointly learns a classifier for source/web domain and transfers the knowledge to target domain at \emph{image-level}. However, for object detection, we need to go deeper to \emph{instance-level}. In particular, we need to learn $f:{x}^w, {x}^t \rightarrow {Y}$. Therefore, we will need a backbone structure to learn from image-level labels and propagate knowledge to instances, and an effective way to transfer knowledge from the web domain to the target domain at instance-level.

\section{Methodology}
\label{method}
Fig.~\ref{fig:network_architecture} shows the diagram of the proposed framework for zero-annotation object detection with web knowledge transfer. The entire framework branches into three streams after feature representation, including WSD, DA, and ST. 
In the following, we describe each stream one by one.

\subsection{Weakly supervised detection trained on web images}\label{sec:wsd}
Our weakly supervised detection backbone is based on the basic WSDDN~\cite{bilen2016weakly} (blue region in Fig.~\ref{fig:network_architecture}). Note that other end-to-end WSD methods can be easily applied as well. Specifically, for WSSDN, the proposal features ${x}_{i} \in {\mathbbm{R}^{d}}$ are obtained through an ROI pooling layer on the feature map of the image, followed by two fully connected layers, similar to Fast-RCNN~\cite{girshick2015fast}.  Then we represent each image $X$ as the concatenation of its proposal features, i.e., $X = \text{concat}({x}_{i}), \forall i \in \left[1, m\right]$, thus $X \in {\mathbbm{R}^{m \times d}}$, where $m$ denotes the number of proposals in the image. Note that here we abuse the notation ${x}_{i}$ to represent both proposal and its corresponding feature, and $X$ to represent both image and its corresponding concatenated feature matrix.

Following the proposal feature learning, the WSD network breaks into two branches of fully connected ($fc$) layers to produce two score matrices ${S}^{cls}$ and ${S}^{loc} \in{\mathbbm{R}^{m \times C}}$, where $C$ is the number of object classes. Then ${S}^{cls}$ and ${S}^{loc}$ are passed to two $softmax$ layers with different axes, i.e. ${S}^{cls}$ is normalized in the class dimension to produce the class probability of each proposal and ${S}^{loc}$ is normalized in the proposal dimension to find the most responsive proposal for each class among all candidate proposals. For proposal $i$ and class $c$, we respectively denote the outputs of these two $softmax$ layers as $p^{cls}_{i,c}$ and $p^{loc}_{i,c}$, which are defined as
\begin{equation}\label{eq:wsd1}
p^{cls}_{i,c} = \dfrac{e^{{s}^{cls}_{i,c}}}{\sum_{k=1}^{C}e^{{s}^{cls}_{i,k}}} {,}\quad
p^{loc}_{i,c} = \dfrac{e^{{s}^{loc}_{i,c}}}{\sum_{k=1}^{m}e^{{s}^{loc}_{k,c}}}
\end{equation}

Then the detection probability $p_{i,c}$ of each proposal can be computed by element-wise products of the normalized probabilities from the two branches:
\begin{equation}\label{eq:wsd2}
p_{i,c} = p^{cls}_{i,c} \cdot p^{loc}_{i,c}.
\end{equation}
The image classification probability $p_{c}$ is calculated by summing up the detection probabilities of all proposals:
\begin{equation}\label{eq:wsd3}
p_{c} = \sum_{i=1}^{m} p_{i,c}.
\end{equation}
Finally, the multi-class cross entropy loss is adopted as the loss function of WSD, which is defined as
\begin{equation}\label{eq:loss_wsd}
L_{WSD} = -\sum_{c=1}^{C} [ y(c) log(p_{c}) + (1 - y(c)) log(1 - p_{c}) ]
\end{equation}
where $y(c) \in\{0,1\}$ is the web image label for class $c$.

Note that since we do not have any label in the target domain, this WSD loss is only optimized by training with web images.

\subsection{Instance-level adversarial domain adaptation}\label{sec:domain_adaptation}
\begin{figure}[t]
\begin{center}
 \includegraphics[width=0.6\linewidth]{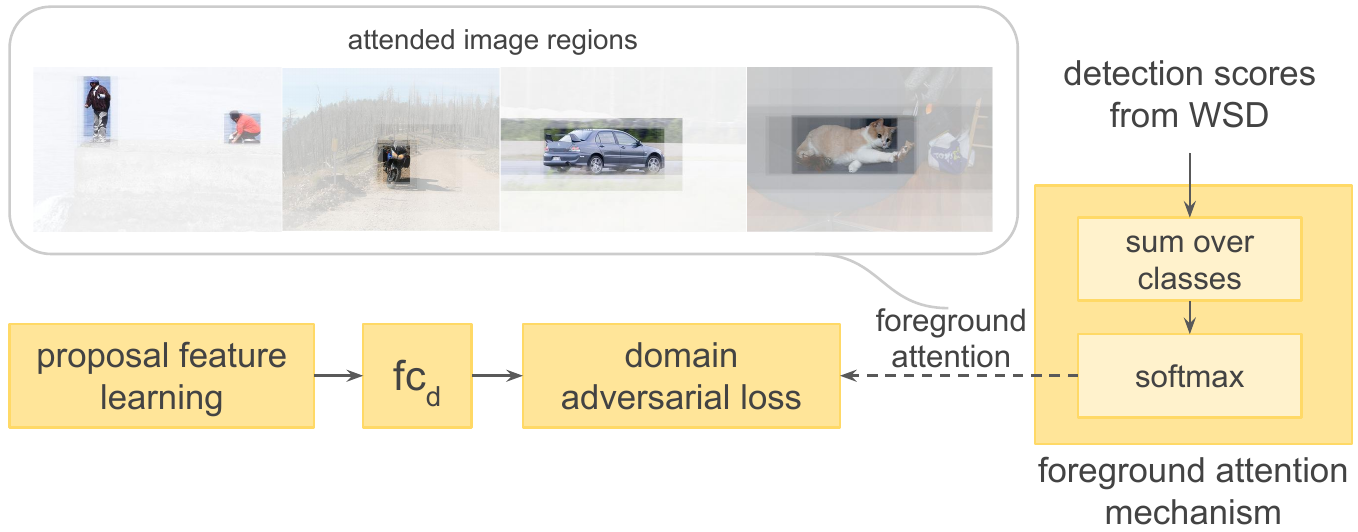}
\end{center}
  \caption{Instance-level domain adaptation stream with foreground attention. We visualize the attended image regions produced by the foreground attention mechanism. The examples show that the foreground object regions are well attended and the background regions are suppressed during domain adversarial learning.}\label{fig:da}
\end{figure}

The purpose of this instance-level domain adaptation (DA) stream (yellow region in Fig.~\ref{fig:network_architecture}) is to close the feature discrepancies between the two domains. Fig.~\ref{fig:da} gives the detailed structure of this DA stream. In particular, it includes two players with adversarial goals: a discriminator trained to differentiate the domains where input features come from, and a feature learner shared with the WSD stream trained to align features from both domains so as to confuse the discriminator.

In particular, the proposed discriminator consists of a fully connected layer $fc_d$ that classifies the input proposal features $x_i$ in $i$-th row of $X$ to their domains $y^t_i \in\{0,1\}$. Here we define $y^t_i = 0$ for $x_i$ from the web domain $D^{w}$ and $y^t_i = 1$ for $x_i$ from target domain $D^{t}$. Through a $softmax$ operation, we can compute the domain probability as $p^t_i$, i.e $prob(y^t_i=1)$. The adversarial loss can then be written as
\begin{equation}
\label{eq:loss_da}
\centering
\begin{split}
\min_{\phi_{f}^w}\max_{\phi_{fc_d}}
\mathbbm{E}_{{x}\sim D^t} [\log(p^t) ]
+ \mathbbm{E}_{{x}\sim D^w} [\log(1-p^t)], \\
\mathbbm{E}_{{x}\sim D^t} [\log(p^t)] = \sum_{i}\mathbbm{1}[y^t_i=1]log(p^t_i), \\
\mathbbm{E}_{{x}\sim D^w} [\log(1-p^t)] = \sum_{i}\mathbbm{1}[y^t_i=0]\log(1-p^t_i),
\end{split}
\end{equation}
where $\phi_{f}^w$ denotes the parameters of the feature learner, $\phi_{fc_d}$ denotes the parameters of the discriminator $fc_d$, and $\mathbbm{1}[]$ is the indication function.

The optimization of the minimax domain adversarial loss in~\eqref{eq:loss_da} is achieved by training alternatively between the following two steps. First, we update $\phi_{fc_d}$ to distinguish proposal features from $D^w$ and $D^t$ to seek for maximizing the loss. Then we fix $\phi_{fc_d}$ and learn the feature representation $\phi_{f}^w$ to minimize the loss so as to confuse the discriminator. In practice, we only shift the web domain $D^w$ towards the target domain and $\phi_{f}^w$ is updated by training only web images.

Moreover, unlike the existing domain adaptation works for image classification~\cite{tzeng2015simultaneous,ganin2015unsupervised}, which focus on aligning image-level features, here we need to align instance-level features instead, especially for important instances that are more likely to contain objects. Specifically, while adapting the instance-level features, we care more about the foreground features than those background features in order to learn common object appearances. Therefore, we introduce an attention mechanism to focus on the adaptation of foreground features and suppress the effects for background features. As shown in Fig.~\ref{fig:da}, our foreground attention model uses the detection scores from the WSD stream and computes the foreground probability $p^f_i$ for proposal $i$ by summing up $p_{i,c}$ in~\eqref{eq:wsd2} over all the classes (i.e. $\sum_{c=1}^{C} p_{i,c}$) followed by a $softmax$ operation for the normalization over all the proposals. This is to find out the most responsive proposals regardless which object classes they belong to, and the responsive proposals with high $p^f$ scores are highly likely to be foreground. Finally, we use the foreground probability as the attention weight, and modify the minimax adversarial loss as
\begin{equation}
\label{eq:loss_da_fa}
\min_{\phi_{f}^w}\max_{\phi_{fc_d}}
\mathbbm{E}_{{x}\sim D^t} [p^f \cdot \log(p^t)] 
+ \mathbbm{E}_{{x}\sim D^w} [p^f \cdot\log(1-p^t)].   
\end{equation}

\subsection{Simultaneous transfer by pseudo supervision}\label{sec:pseudo_sup}

\begin{figure}[t]
\begin{center}

 \includegraphics[width=0.6\linewidth]{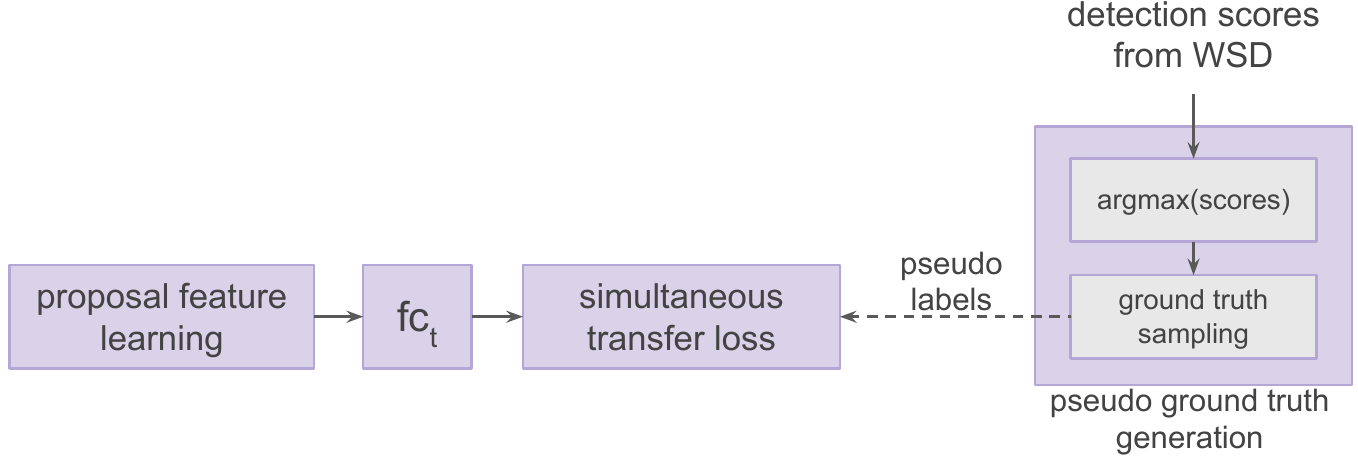}
\end{center}
  \caption{Simultaneous transfer stream with pseudo ground truth generation.}\label{fig:st}
\end{figure}

Ideally, the domain adaptation stream should produce domain invariant features and improve the detection results while applying on the target dataset. However, it is observed that it fails to perform domain shift with class-specific directions. Specifically, it could encourage the features to be indistinguishable across not only domains but also classes. 
This ill effect of DA stream eventually makes features to be non-discriminative. Therefore, to preserve the semantic structure across different categories, we introduce the simultaneous transfer (ST) stream (purple part in Fig.~\ref{fig:network_architecture}) and use the pseudo labels generated from the WSD network as the supervision to preserve or even enhance the discriminative power of the learned features. The network details are shown in Fig~\ref{fig:st}.

To generate the pseudo ground truth for each target image, we use the detection scores $p_{i,c}$ in~\eqref{eq:wsd2} from the WSD stream. We select to highest scoring proposal for each object class c, denoted as $i_{c} = \argmax_i p_{i,c}$. We set a threshold $t$ to determine the presence of a class in an image. If $p_{i_c,c} >= t$, the corresponding proposal $i_{c}$ is selected as the pseudo ground truth bounding box. Given the pseudo ground truth boxes, we then sample the boxes with large overlaps with the pseudo ground truth boxes as positive examples and randomly sample a few background examples from the remaining bounding boxes. 

Finally, we use the $softmaxloss$ as the ST loss function:
\begin{equation}\label{eq:loss_st}
L_{ST} =-\sum_{i\in{P}}\sum_{c=0}^{C}\mathbbm{1}[y^{ST}_i = c]log(p^{ST}_{i,c}),
\end{equation}
where $y^{ST}\in{\{0,1,2,...C\}}$ are the class labels  (0 is the class label of background), $P$ is the set of the selected proposals, and $p^{ST}_{i,c}$ is the class probability output from the fully connected layer $fc_t$ followed by a $softmax$ operation.  

Conditional adversarial loss is also a common way in GAN to enable class-specific domain adaptation. However, here the conditions are instance-level pseudo labels, which are noisy labels. It will be more stable to detach the class conditional learning with the domain adversarial learning. 

\section{Experiments}
In this section, we conduct various experiments to evaluate the effectiveness of our proposed zero-annotation object detection  with web knowledge transfer.

\subsection{Datasets and experiment setup}
We evaluate our method on two object detection benchmark datasets: Pascal VOC 2007 and 2012. These two datasets contain images of 20 object classes.  The web images we used are from the STC dataset~\cite{wei2017stc}, whose images can be freely obtained from Internet without human labour. 
Similar as most supervised detection works, mean average precision (mAP) is used as the evaluation metric. Following the common standard, the IoU threshold is set to be 0.5 between ground truths and correctly predicted boxes.


\textbf{Implementation details.}
Our method is built upon two pre-trained networks on imagenet: VGG\_M and VGG 16.  We use selective search~\cite{uijlings2013selective} to generate proposals for source and target images. 
In the WSD stream, we follow the details in the basic model of WSDDN as described in Section~\ref{sec:wsd}. The ROI features from the web domain are passed to the WSD stream to optimized the WSD loss whereas the ROI features from the target domain are only forwarded up to the detection score layer to generate foreground attention weights for the DA stream and pseudo labels for the ST stream. 
The DA stream takes the inputs from both source and target domains. It alternates between training the discriminator and the feature generator each time after training 5000 images. Lastly, the ST stream takes the inputs from the target domain and uses the detection scores generated from the WSD stream to generate pseudo ground truths as described in Section \ref{sec:pseudo_sup}.

\subsection{Baseline and upper bound}
\begin{table}
\begin{center}
\caption{Baseline(wt.web data) and upper-bound(wt.VOC labels) on VOC 2007.}\label{tb:baseline_upperbound}
\centering
\begin{tabular}{|l|c|}
\hline
Method & mAP \\
\hline\hline
WSD(wt.web data)-VGG\_M & 21.5 \\
WSD(wt.web data)-VGG16 & 21.8 \\
\hline
WSD(wt.VOC labels)-VGG\_M &30.2\\
WSD(wt.VOC labels)-VGG16 &29.3\\
\hline
\end{tabular}
\end{center}
\end{table}

The baseline of our method is the basic WSD network~\cite{bilen2016weakly} trained using only web images with web image labels. As shown in Table \ref{tb:baseline_upperbound}, due to the domain mismatch,  the results are only 21.5 for VGG\_M and 21.8 for VGG16. 

The upper bound of our method is to train the basic WSD network with VOC image-level labels similar as~\cite{bilen2016weakly}. Our obtained upper bound result for VGG\_M is quite close to that reported in~\cite{bilen2016weakly} with selective search proposals, while our result of 29.3 for VGG16 is higher than that of 24.3 reported in~\cite{bilen2016weakly} with selective search proposals. Also, we have the same finding as~\cite{bilen2016weakly} that VGG 16 performs slightly worse than VGG\_M. This could be because the image level labels might not give sufficient supervision for a very deep network for the MIL problem.

Overall, there are significant gaps between the results without VOC labels and those with VOC labels. We aim to reduce the gap between the unsupervised and weakly supervised detection by transferring the knowledge of web domain to target domain with our proposed method.

\subsection{Detailed results and analysis}\label{sec:results_compared_with_baseline}

Table~\ref{tb:results_2007} shows the detailed detection results of different combinations of the three streams developed in our method on VOC2007 test set. All of these methods are evaluated against the baseline,`WSD(Baseline)', that uses web images to train the WSD network alone. Before training the DA and ST streams, we train the WSD for one epoch first. This will give a more stable initialization to get the foreground attention weights for DA and pseudo labels for ST.

\begin{table*}[t]
\setlength\tabcolsep{4pt}
\centering
\caption{Average precision results (\%) of different component combinations on VOC2007 test set.}\label{tb:results_2007}
\resizebox{\textwidth}{!}{%
\begin{tabular}{|c|cccccccccccccccccccc||c| }
\hline
&aero&bike&bird&boat&bottle&bus&car&cat&chair&cow&table&dog&horse&mbike&person&plant&sheep&sofa&train&tv&mean\\ \hline
\hline
WSD(Baseline)-VGG\_M&31.1&27.1&\textbf{18.6}&10.0&9.1&29.9&37.7&21.5&2.7&15.8&21.5&\textbf{27.8}&30.0&35.7&\textbf{10.8}&9.9&17.6&28.9&23.1&21.1&21.5\\
WSD+DA-VGG\_M&30.3&24.1&15.6&\textbf{13.8}&9.1&32.7&\textbf{39.0}&21.4&2.9&19.0&26.4&25.5&24.7&32.9&4.3&8.2&15.6&28.7&24.5&25.1&21.2\\
WSD+DA+ST-VGG\_M&33.3&\textbf{31.5}&16.9&\textbf{13.8}&\textbf{10.8}&39.5&36.2&\textbf{30.8}&8&19.9&33.4&18.4&26.4&37.8&8.3&\textbf{13.1}&15.5&32.1&25.0&33.8&24.2\\
WSD+DA+2ST-VGG\_M&34.3&31.3&18.5&9.4&10.6&39.6&37.7&17.9&\textbf{10.2}&16.7&\textbf{34.7}&19.8&\textbf{31.8}&40.7&7.4&12.5&18.6&\textbf{33.0}&26.8&34.6&24.3\\
WSD+DA+3ST-VGG\_M&\textbf{35.6}&31.3&18.2&7.7&9.1&\textbf{40.4}&38.4&23.8&9.7&\textbf{20.1}&33.4&22.5&30.9&\textbf{41.4}&9.8&10.8&\textbf{18.7}&28.7&\textbf{27.1}&\textbf{34.7}&\textbf{24.6}\\
\hline
WSD(Baseline)-VGG16&\textbf{45.8}&28.2&11.1&8.5&2.5&42.8&\textbf{41.5}&25.9&4.2&15.9&13.0&16.9&\textbf{28.0}&40.8&3.6&5.5&11.0&38.5&28.4&23.2&21.8\\
WSD+DA-VGG16&33.8&22.4&13.1&13.4&9.1&38.1&36.5&25.8&9.2&20.1&12.6&\textbf{19.8}&19.9&34.4&4.4&10.8&13.8&30&26.8&25.1&21.0\\
WSD+DA+ST-VGG16&43.7&30.8&15.7&10.6&\textbf{13.4}&41.3&39.5&23.9&\textbf{12.8}&\textbf{20.7}&27.9&13.9&23.4&39.7&\textbf{10.3}&12.7&\textbf{21.3}&\textbf{39.6}&28.1&30.7&25.0\\
WSD+DA+2ST-VGG16&44.7&\textbf{31.0}&12.1&15.7&11.8&38.8&40.6&29.1&12.0&17.9&\textbf{32.2}&9.1&24.1&42.8&7.6&\textbf{13.7}&17.0&33.4&30.6&\textbf{33.5}&24.9\\
WSD+DA+3ST-VGG16&40.6&30.1&\textbf{17.8}&\textbf{15.9}&6.4&\textbf{42.9}&40.5&\textbf{31.5}&11.4&20.3&27.4&15.7&24.1&\textbf{43.8}&8.9&12.2&17.7&37.3&\textbf{32.1}&31.0&\textbf{25.4}\\
\hline
\end{tabular}
}
\end{table*}

\begin{table*}[t]
\setlength\tabcolsep{4pt}
\centering
\caption{Average precision results (\%) on VOC2012 test set.}\label{tb:results_2012}
\resizebox{\textwidth}{!}{%
\begin{tabular}{|c|cccccccccccccccccccc||c| }
\hline
&aero&bike&bird&boat&bottle&bus&car&cat&chair&cow&table&dog&horse&mbike&person&plant&sheep&sofa&train&tv&mean\\ \hline
\hline
WSD(Baseline)-VGG\_M&39.7&25.4&12.6&5.8&2.3&32.3&\textbf{25.0}&20.7&1.6&17.9&9.6&\textbf{29.0}&24.3&42.4&3.8&4.6&10.6&16.6&22.5&11.4&17.9\\
WSD+DA+3ST-VGG\_M&\textbf{44.3}&\textbf{29.8}&\textbf{15.6}&\textbf{6.6}&\textbf{6.0}&\textbf{34.4}&24.2&\textbf{25.1}&\textbf{5.7}&\textbf{20.3}&\textbf{22.3}&24.9&\textbf{29.1}&\textbf{45.2}&\textbf{7.8}&\textbf{9.4}&\textbf{12.4}&\textbf{21.4}&\textbf{22.6}&\textbf{26.0}&\textbf{21.7}\\
\hline
WSD(Baseline)-VGG16&47.9&29.2&14.8&\textbf{7.9}&3.5&\textbf{39.6}&\textbf{27.3}&24.6&2.3&15.9&4.9&18.3&25.5&\textbf{47.5}&3.8&4.3&9.4&22.2&19.3&16.0&19.2\\
WSD+DA+3ST-VGG16&\textbf{48.8}&\textbf{32.8}&\textbf{16.6}&6.3&\textbf{7.7}&39.0&26.2&\textbf{32.6}&\textbf{7.8}&\textbf{18.3}&\textbf{12.4}&\textbf{22.1}&\textbf{29.7}&45.9&\textbf{9.6}&\textbf{9.0}&\textbf{14.5}&\textbf{24.0}&\textbf{26.8}&\textbf{28.1}&\textbf{22.9}\\
\hline
\end{tabular}
}
\end{table*}

From Table~\ref{tb:results_2007}, we can see that adding DA alone, `WSD+DA', results in a slight drop in mAP. As discussed in Section~\ref{sec:domain_adaptation}, DA could result in an unexpected feature confusion among object classes with similar appearance, such as vehicle classes and animal classes. Only for classes that are different from all the other classes, such as ``tv monitor", DA shows its contribution to the detection results. 

It can be seen that by further adding the ST stream, `WSD+DA+ST', the detection results improve significantly, by 2.7\% for VGG\_M and 3.2\% for VGG16, compared with the baselines. In addition, inspired by the idea of~\cite{tang2017multiple}, we also evaluate the performance of adding multiple pseudo label transfer streams one by one. Specifically, the pseudo labels generated by the first ST stream are used as the supervision of the second ST stream, whose generated pseudo labels are then used as the supervision of the third ST stream. The results of appending multiple ST streams, `WSD+DA+2ST' and `WSD+DA+3ST', are also shown in Table~\ref{tb:results_2007}. We can see that adding one additional ST stream generally leads to slight improvements. Overall, by adding the ST streams, our method brings up the results for most categories, especially difficult classes such as ``chairs" and ``dining tables". These classes are usually in cluttered scenes and the single WSD learned from clean web images can hardly capture the objects from the environment.

The overall performance gains from the best combinations are 3.1\% for VGG\_M and 3.6\% for VGG16. These results show that our proposed method improves the baseline webly supervised detection model significantly by introducing the DA and ST streams. In VGG16, it brings up the unsupervised results to 25.4\% without any labels from the target dataset, much closer to the weakly supervised result of 29.3\% that requires image-level labels from the target dataset.

In addition to the mAP results for detection, we also measured the correct localization (CorLoc) result on the VOC 2007 trainval set (see Table. \ref{tb:coloc} ) and compare it with the best reported CorLoc results of the WSD works \cite{bilen2015weakly,bilen2016weakly,tang2017multiple}. Note that all of these WSD methods use image labels of VOC trainval set during the training and CorLoc is measured on these training images. In our method, we do not include any VOC training labels and we can still achieve a good localization model, 44.3\% images are with correctly localized objects, which is even better than \cite{bilen2015weakly}.

\subsection{Ablation experiments}\label{sec:ablation_results}
In the following sections, we analyze the effectiveness of each component, including the domain adaptation stream and the simultaneous transfer streams.

\subsubsection{Analysis of the DA stream.} 
\begin{figure*}[t]
\begin{center}
\includegraphics[width=1.0\linewidth]{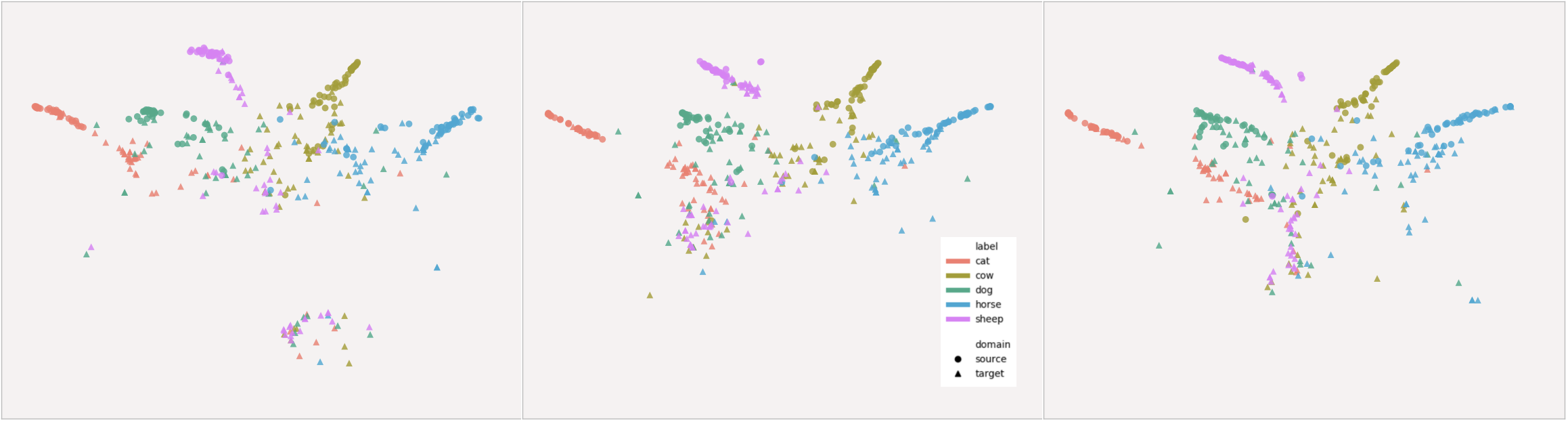}
\end{center}
  \caption{Visualization of features in 2D space by t-SNE~\cite{maaten2008visualizing}. We randomly sample some object proposals from target and web domains and extract fc7 features (VGG\_M) using different methods. Then we use PCA and t-SNE to reduce the dimension to 2. We plot the scatter diagrams for all mammal animal classes. Left: WSD (baseline). Middle: WSD+DA. Right: WSD+DA+3ST.}\label{fig:domain_2d_vis}
\end{figure*}

To further verify the effects of the DA stream, we visualizing the feature distributions of `WSD+DA' in 2D space by t-SNE~\cite{maaten2008visualizing} in Fig.~\ref{fig:domain_2d_vis}. Although this visualization of high dimensional features in 2D space may not be accurate, we can still have some ideas that the DA stream does help shift the features closer to the same region across domains. 

We further examine the results by removing the DA stream from the overall structure. As shown in Table~\ref{tb:da_eval}, the WSD with the ST streams only  cannot achieve as high detection mAP as our overall network with the DA stream, which demonstrates the contribution of DA to the overall network. In~Table~\ref{tb:da_eval}, we also evaluate the effectiveness of the foreground attention mechanism (FA) for the DA stream. It can be seen that the result of DA without FA, `WSD+DA(w/o.FA)+3ST', is even worse than of no DA, `WSD+3ST', which suggests that treating all proposals equally during DA does not help.

\begin{table}[!htb]
\begin{minipage}{.45\linewidth}
\caption{Comparing the results (mAP in \%) on VOC 2007 test set with different settings of the DA stream.}\label{tb:da_eval}
\begin{tabular}{|l|c|}
\hline
Method & mAP \\
\hline\hline
WSD+3ST-VGG\_M & 23.5 \\
WSD+DA(w/o.FA)+3ST-VGG\_M & 23.3 \\
WSD+DA+3ST-VGG\_M & \textbf{24.6}\\
\hline
\end{tabular}
\end{minipage}%
\hspace{.1\linewidth}
\begin{minipage}{.45\linewidth}
\centering
\caption{CorLoc results on VOC 2007 compared with WSD methods.}\label{tb:coloc}
\begin{tabular}{|l|c|}
\hline
Method & CorLoc \\
\hline\hline
Bilen et al \cite{bilen2015weakly} & 43.7 \\
Bilen et al \cite{bilen2016weakly} & 56.1 \\
Tang et al \cite{tang2017multiple} &60.6\\
WSD+DA+3ST-VGG16 &44.3\\
\hline
\end{tabular}
\end{minipage} 
\end{table}

\subsubsection{Analysis of the ST stream.} We also visualized the features of WSD+DA+3ST in 2D space in Fig.~\ref{fig:domain_2d_vis}. It can be seen that by adding both DA and ST, we are able to move the cross-domain features closer while making the classes in target domain more separable.

We would like to point out that the incremental gains of our method with multiple ST streams are not as much as~\cite{tang2017multiple} that also use multiple refinement streams. This is due to the following reason. In~\cite{tang2017multiple}, the positive samples are selected by image-level labels of target dataset and their purpose is to refine the instance classifier for multiple times. However, our method does not use the image labels of VOC dataset and our purpose is to prevent the unexpected distribution shift among similar classes. In other words, the gain of pseudo label transfer in our scenario is mainly from the effects of preserving the semantic structure among classes rather than refining the instance classifiers again and again.

One insight of the ST stream is that our framework trains the WSD model from web domain and selects pseudo ground truth samples of target domain based on the current WSD model at the same time. In other words, the ST stream is trained simultaneously with the WSD stream. In this way, it shares the feature learning between the WSD stream for web image training and the ST stream for target dataset training. An alternative way of transferring the pseudo labels is to train on the two datasets in an isolated way. In particular, we can first pre-train the WSD using web images, then use this pre-trained WSD model to generate the pseudo ground truths for the target dataset and finally use these pseudo ground truths to train a detector for target dataset. We conduct the experiment using such isolated method and obtain an mAP of 22.5\%. This implies that the simultaneous weights sharing is important for the learning transfer across domains.

\subsection{More results} We also evaluate our method on VOC 2012 dataset and the results are shown in Table \ref{tb:results_2012}. The baseline result shows that the detection model trained using only web images gives poor results for VOC 2012 test images. By adding our DA stream and ST streams, the results are largely improved for most classes. Overall, we achieve significant increases of 3.8\% and 3.7\% in mAP with VGG\_M and VGG16 respectively for VOC 2012 dataset. 

\section{Conclusion}
In conclusion, we introduced an annotation-free object detection method by learning from web image resources. Particularly, to solve the domain mismatch problem between the web domain objects and the target domain objects, we proposed an instance-level domain adaptation stream with foreground attention, together with a simultaneous transfer stream that simultaneously learns target data from pseudo labels. Through these novel components, we achieved significant improvements in detection results and successfully reduced the performance gap between the baseline detectors trained with and without human annotations.

\textbf{Acknowledgements.}This project is partially supported by MoE Tier-2 Grant (MOE2016-T2-2-065).
%
%
%
\bibliographystyle{splncs04}
\bibliography{egbib}
%




\end{document}